\documentclass[10pt,twocolumn,letterpaper]{article}

\usepackage[pagenumbers]{cvpr} 
\usepackage{overpic}
\usepackage{makecell}
\usepackage{adjustbox} 
\usepackage{multirow}
\usepackage{float}
\usepackage{bbding}
\usepackage{overpic}
\usepackage{times}
\usepackage{epsfig}
\usepackage{graphicx}
\usepackage{amsmath}
\usepackage{amssymb}
\usepackage{booktabs}
\usepackage{xcolor}
\usepackage{enumitem}
\usepackage{lipsum}
\usepackage{diagbox}

\usepackage[ruled,vlined,linesnumbered]{algorithm2e}
\makeatletter
\newcommand{\algorithmfootnote}[2][\footnotesize]{%
  \let\old@algocf@finish\@algocf@finish
  \def\@algocf@finish{\old@algocf@finish
    \leavevmode\rlap{\begin{minipage}{\linewidth}
    #1#2
    \end{minipage}}%
  }%
}
\makeatother

\definecolor{citecolor}{HTML}{0071bc}
\usepackage[pagebackref=false,breaklinks=true,letterpaper=true,colorlinks,citecolor=citecolor,bookmarks=false]{hyperref}

\newcommand{\app}{\raise.17ex\hbox{$\scriptstyle\sim$}}

\definecolor{codegreen}{rgb}{0.0,0.6,0.0}

\definecolor{cvprblue}{rgb}{0.21,0.49,0.74}

\def\confName{CVPR}
\def\confYear{2024}

\title{UltimateDO: An Efficient Framework to Marry 
Occupancy Prediction
with 
3D Object Detection 
via Channel2height
}

\author{
Zichen Yu$^{1}$ \quad Changyong Shu$^{2}$$\textsuperscript{\Envelope}$ \\
$^1$Dalian University of Technology, $^2$Houmo AI \\
{\tt\small yuzichen@mail.dlut.edu.cn} {\tt\small changyong.shu@houmo.ai}
\vspace{-15pt}
}

\begin{document}

\maketitle
\begin{abstract}

Occupancy and 3D object detection are characterized as two standard tasks in modern autonomous driving system.
In order to deploy them on a series of edge chips with better precision and time-consuming trade-off,
contemporary approaches either deploy standalone models for individual tasks, or design a multi-task paradigm with separate heads. 
However, they might suffer from deployment difficulties (i.e., 3D convolution, transformer and so on) or deficiencies in task coordination. 
Instead, we argue that a favorable framework should be devised in
pursuit of ease deployment on diverse chips and high precision with little time-consuming. 
Oriented at this, we revisit the paradigm for interaction between 3D object detection and occupancy prediction, reformulate the model with 2D convolution and prioritize the tasks such that each contributes to other. 
Thus, we propose a method to achieve fast 3D object detection and occupancy prediction (UltimateDO),
wherein the light occupancy prediction head in FlashOcc is married to 3D object detection network, with negligible additional time-consuming of only 1.1ms while facilitating each other. 
We instantiate UltimateDO on the challenging nuScenes-series benchmarks. 

\end{abstract}
\section{Introduction}\label{sec1}

\begin{figure}
\centering
		\includegraphics[width=1.0\linewidth]{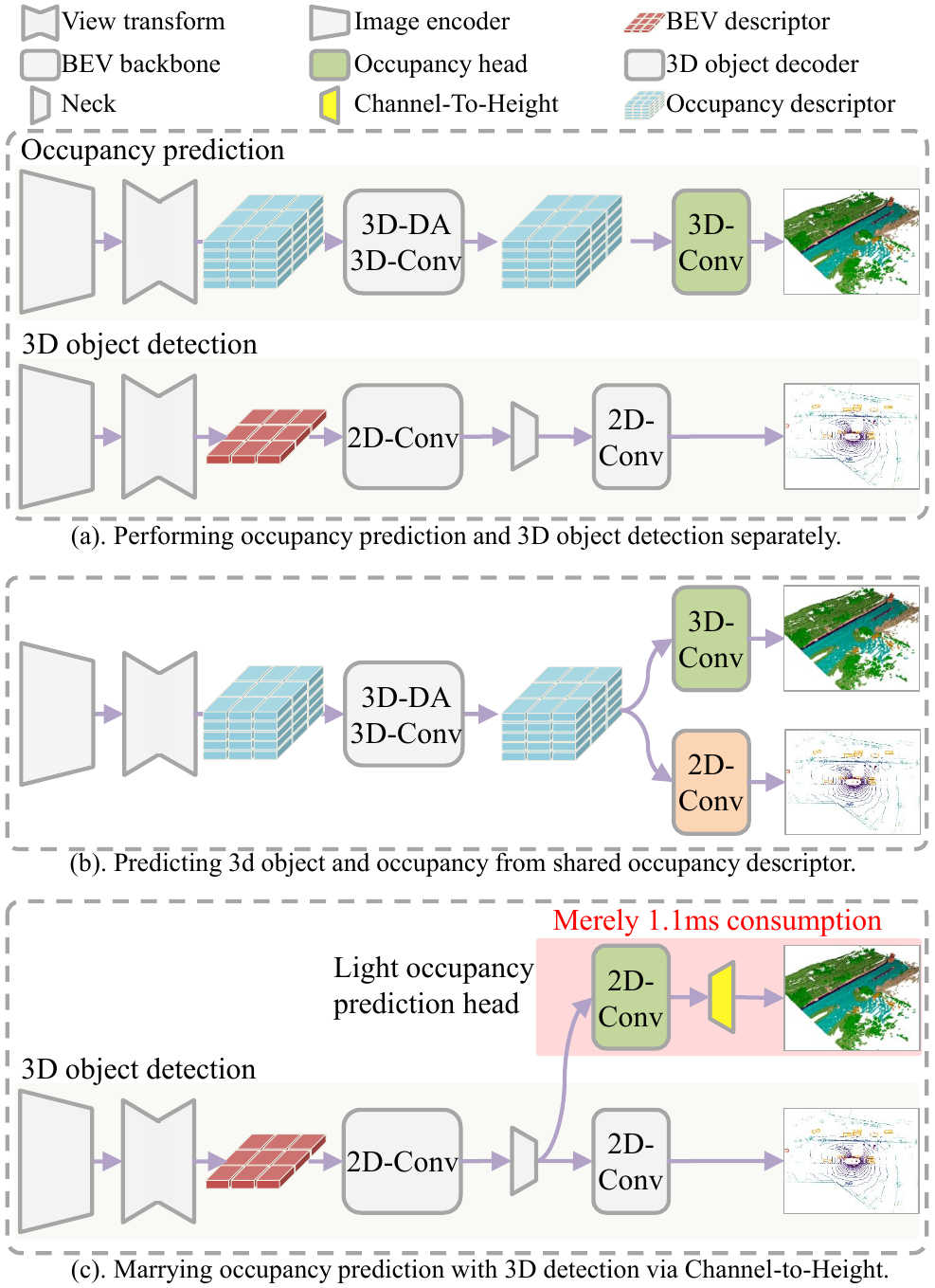}

	\caption{
    \textbf{Different paradigm of interaction between 3D object detection and occupancy prediction}.
    (a) performs occupancy prediction and 3D object detection separately.
    (b) shows the synchronous perception for detection and occupancy from shared voxel-level occupancy descriptor.
    (c) exhibits our UltimateDO where a light grafting occupancy module is married to 3D object detection.
    The abbreviation "Conv" represents convolution, "DA" denotes deformable convolution.
    Besides, the presence of "3D-DA", "3D-Conv" or "2D-Conv" in the icon indicates that the corresponding module is composed of these operators.    
    Best viewed in color.
    }

	\label{fig:fig1}
\end{figure}

Holistic perception, encompassing both object-level and voxel-level representations, plays a crucial role in autonomous driving.
In particular, the acquisition of 3D object bounding boxes and accurate occupancy predictions has gained significant traction among scholars and industry professionals alike, owing to their fundamental significance in ensuring the safety and dependability of autonomous vehicles.
Early studies have traditionally employed separate networks to model object-level and voxel-level understanding. While this divide-and-conquer strategy simplifies algorithm development\cite{zhang2023occformer,pan2023renderocc,tian2023occ3d,wang2023panoocc,huang2021bevdet,petr,petrv2,shu20233DPPE,bevformer}, it comes at the expense of sacrificing the jointly optimized enhancement for each task.

With the emergence of the end-to-end unified solution for autonomous driving system\cite{hu2023_uniad}, an increasing number of studies have attempted to jointly integrate occupancy and detection within the same model. \cite{Occupancy3ddet,occ-bev} inject the occupancy knowledge to help more accurate 3D detection, due to the fine-grained voxel-level semantic information can offer enhanced global perception capabilities. 
Certainly, the incorporation of instance-level discrimination within object detection can significantly enhance the distinctiveness of occupied voxels associated with different semantic labels. Consequently, this augmentation contributes to bolstering the certainty and accuracy in voxel-level semantic classification. 
Hence, the construction of a shared three-dimensional occupancy descriptor is proposed in \cite{tian2023occ3d,pan2023renderocc}, enabling the unified execution of 3D object detection and occupancy prediction tasks.

However, the procession of three-dimensional voxel-level representations inevitably introduces complex computations, such as 3D (deformable) convolutions, transformer operators, and so on\cite{zhang2023occformer,pan2023renderocc,sima2023_occnet,tian2023occ3d,occ-bev,Occupancy3ddet}. 
This poses significant challenges for on-chip deployment and computational power requirements. Sparse occupancy representation is investigated in \cite{wang2023panoocc} to conserve memory resources.
However, this approach does not fundamentally address the challenges for deployment and computation.

Noting the recent advancements in the utilization of FlashOcc~\cite{yu2023flashocc} for BEVDetOcc~\cite{bevdetocc}, wherein all 3D convolutions are replaced with their 2D counterparts and a channel-to-height plugin is introduced.
We observe that the main components of both BEVDet and FO(BEVDetOcc) remain consistent, the only differing components are the occupancy prediction head in FO(BEVDetOcc) and the detection head in BEVDet.
This observation serves as a motivation to marry the occupancy prediction head to BEV neck of 3D object detection model, termed as UltimateDO.
Consequently, our proposed UltimateDO successfully accomplishes the simultaneous execution of occupancy and detection tasks, while facilitating mutual enhancement between them.
Extensive experiments reveals that the UltimateDO framework attains state-of-the-art (SOTA) performance with
negligible additional time-consuming of only 1.1ms for the grafted occupancy prediction head.
Additionally, our framework demonstrates superior performance on the nuScenes dataset.

\section{Related Work}

\textbf{Image-based 3D Object Detection.}
Image-based 3D object detection has become a crucial element within autonomous driving perception systems, owing to its cost-effectiveness.
It garners significant interest from both academia and industry. 
Previous approaches predominantly employed the paradigm of monocular 3D obstacle detection to perceive targets in the image perspective space. Subsequently, the perception results from multiple surround-view cameras were fused during the post-processing stage.
Recently, there has been a notable shift towards employing the Lift-Splat-Shoot (LSS) technique to project surround-view image features onto the Bird's Eye View (BEV) space for subsequent object detection in the BEV domain~\cite{huang2021bevdet,li2022bevdepth}. 
Concurrently, the DETR-like paradigm has also gained significant attention and exploration. Specifically, it can be further classified into two categories: 3D-to-2D~\cite{liu2022petr,wang2023exploring,shu20233DPPE} and 2D-to-3D approaches~\cite{wang2022detr3d,li2022bevformer}.
However, the results of obstacle detection are presented as coarse-grained rectangular bounding boxes (missing fine-grained details of object structures). Additionally, directly classifying unseen objects as background is not conducive to downstream planning tasks. Thus, occupancy prediction is proposed to address the above limitations.

\textbf{3D Occupancy Prediction.}
The earliest origins of 3D occupancy prediction can be traced back to Occupancy Grid Maps (OGM)~\cite{thrun2002probabilistic}, which aimed to extract detailed structural information of the 3D scene from images, and facilitating downstream planning and navigation tasks.
The existing studies can be classified into sparse perception and dense perception based on the type of supervision. The sparse perception category obtains direct supervision from lidar point clouds and are evaluated on lidar datasets~\cite{huang2023tri}.
Simultaneously, dense perception shares similarities with semantic scene completion (SSC)~\cite{armeni2017joint,dai2017scannet}.
Voxformer~\cite{li2023voxformer} utilizes 2.5D information to generate candidate queries and then obtains all voxel features via interpolation. Occ3D~\cite{tian2023occ3d} reformulate a coarse-to-fine voxel encoder to construct occupancy representation. RenderOcc~\cite{pan2023renderocc} extract 3D volume feature from surround views via 2D-to-3D network and predict density and label for each voxel with nerf Supervision. Furthermore, several benchmarks with dense occupancy labels are proposed~\cite{tian2023occ3d,sima2023_occnet}.
However, the above methods based on 3D voxel-level representation consumes huge computation, FlashOcc~\cite{yu2023flashocc} utilizes the channel-to-height to perform the occupancy prediction on flattened BEV-level representation, with superior performance while least consumption.

\begin{figure*}
\centering
	\includegraphics[width=1.0\linewidth]{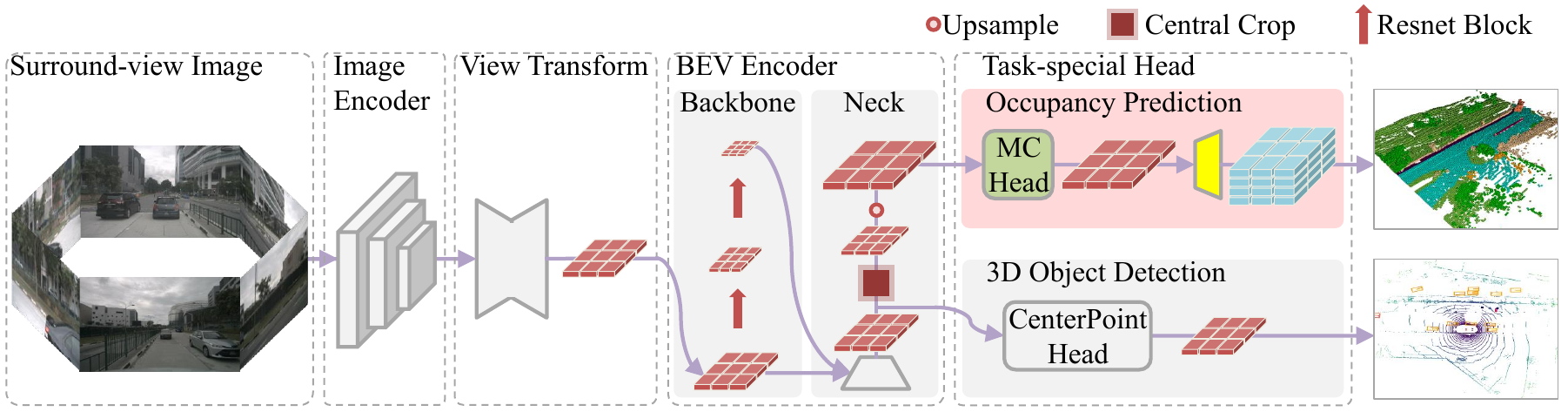}
	\caption{The diagram illustrates the overarching architecture of our proposed UltimateDO, which is best viewed in color and with zoom functionality. The region designated by the dashed box indicates the presence of replaceable modules. The light blue region corresponds to the optional temporal fusion module, and its utilization is contingent upon the activation of the red switch.
    MC is short for multi-convolution.
    Moreover, apart from the instructions provided for the three special icons located in the upper right corner, all remaining icons comply with the guidelines presented in Figure~\ref{fig:fig1}.
	}
	\label{fig:overview}
\end{figure*}

\textbf{Occupancy Interacted with Detection.} 
A comprehensive understanding of 3D scene is beneficial for promoting the performance of the entire perception pipeline.
Fine-grained structured information in occupancy representation facilitates 3D detection, ~\cite{Occupancy3ddet} introduces occupancy attention learning for more accurate detection, 
~\cite{occ-bev} perform 3D detection via initialized from an binary occupancy pretrained model.
However, the above methods cannot achieve detection and occupancy simultaneously.
Existing methods~\cite{sima2023_occnet,wang2023panoocc} aim to integrate detection and segmentation into a unified learning framework, enabling dense scene semantics understanding with object-level instance discrimination, thus occupancy and detection can benefit each other under the joint-training framework.
~\cite{sima2023_occnet} constructs shared occupancy descriptor in a cascade fashion where voxels are refined progressively, which is then directly fed into occupancy and detection. 
~\cite{wang2023panoocc} takes a step further to explore the sparse representation of occupancy to boost memory efficiency. 
The 3D voxel-level representation in these methods necessitates 3D computations, which inevitably leads to increased memory requirements and higher hardware deployment demands compared to purely 2D convolution models.

\section{Framework}
As illustrated in Figure.~\ref{fig:overview}, 
the UltimateDO framework takes surround-view images as input and produces multiple outputs, including the forecasted 3D object bounding box with velocity information and dense occupancy prediction results.

Specifically, 
the surrounding-view images are fed into a 2D image encoder to extract high-level features in the perception-view.
These features then undergo view transformation to obtain flattened BEV representations.
The above coarse BEV information is further refined by passing through the BEV encoder, resulting in enhanced BEV features.
The refined BEV features are utilized by task-specific heads for downstream perception tasks. 

Indeed, considering that the perception range may vary between 3D object detection and occupancy prediction, as explicitly described in the training details of the experiment section, 
a central crop module is introduced to adjust the size of the BEV features specifically for the occupancy prediction branch,
and an upsampling operator is employed to adjust the feature size to match the ground truth occupancy in the BEV dimension
This ensures that the occupancy prediction accurately corresponds to the desired perception range.

\section{Experiment}

In this section, we first detail the benchmark and metrics, as well as the training details for our framework in Section.~\ref{sec:exp_setup}. 
Then, Section.~\ref{sec:sota_exp} presents the main results of our framework with fair comparison to other state-of-the-art methods on 3D object detection and occupancy prediction.
After that, we conduct extensive ablative experiments to investigate the effectiveness of each component in our proposed framework in Section.~\ref{sec:exp_ablation}.

\begin{table*}[t]
\scriptsize
\caption{3D occupancy prediction performance on the Occ3D-nuScenes valuation dataset. 
$\ast$ denotes initialized from the pre-trained FCOS3D backbone. 
Cons. Veh represents construction vehicle and Dri. Sur is for driveable surface.
The frame per second (FPS) metric is evaluated using RTX3090, employing the TensorRT benchmark with FP16 precision.
And the FPS is tested for the total model including occupancy and detection head.
} 
\setlength{\tabcolsep}{0.005\linewidth}

\def\mystrut{\rule{0pt}{1.5\normalbaselineskip}}
\centering
\begin{adjustbox}{width=2.1\columnwidth,center}
\begin{tabular}{l | c c | c | c | r r r r r r r r r r r r r r r r r}
    \toprule[1.5pt]
    Method 
    & \rotatebox{90}{Backbone}
    & \rotatebox{90}{Image size}
    & \rotatebox{90}{\textbf{mIoU}$\uparrow$}  
    & \rotatebox{90}{\textbf{FPS}$\uparrow$} 
    & \rotatebox{90}{\textbf{others}$\uparrow$} 
    & \rotatebox{90}{\textbf{barrier}$\uparrow$}
    & \rotatebox{90}{\textbf{bicycle}$\uparrow$} 
    & \rotatebox{90}{\textbf{bus}$\uparrow$} 
    & \rotatebox{90}{\textbf{car}$\uparrow$} 
    & \rotatebox{90}{\textbf{Cons. Veh}$\uparrow$} 
    & \rotatebox{90}{\textbf{motorcycle}$\uparrow$} 
    & \rotatebox{90}{\textbf{pedestrian}$\uparrow$} 
    & \rotatebox{90}{\textbf{traffic cone}$\uparrow$} 
    & \rotatebox{90}{\textbf{trailer}$\uparrow$} 
    & \rotatebox{90}{\textbf{truck}$\uparrow$} 
    & \rotatebox{90}{\textbf{Dri. Sur}$\uparrow$} 
    & \rotatebox{90}{\textbf{other flat}$\uparrow$} 
    & \rotatebox{90}{\textbf{sidewalk}$\uparrow$} 
    & \rotatebox{90}{\textbf{terrain}$\uparrow$} 
    & \rotatebox{90}{\textbf{manmade}$\uparrow$} 
    & \rotatebox{90}{\textbf{vegetation}$\uparrow$} 
    \\
    \midrule
    BEVDetOcc~\cite{bevdetocc} & R101$\ast$ & 512$\times$1408 & 32.81 & 27.3 & 7.61 & 38.28 & 9.62 & 39.68 & 46.57 & 17.99 & 14.88 & 18.10 & 10.38 & 30.08 & 33.14 & 80.04 & 38.00 & 49.51 & 52.37 & 38.21 & 33.25 \\
       
    FlashOcc~\cite{yu2023flashocc} & R101$\ast$ & 512$\times$1408 & 33.4 & 29.6 & 7.54 & 39.14 & 11.37 & 40.93 & 47.06 & 14.52 & 14.82 & 16.57 & 11.27 & 30.83 & 33.65 & 80.77 & 41.04 & 49.93 & 53.75 & 40.01 & 34.22 \\
       
    \midrule
    \bf{UltimateDO} & R101$\ast$ & 512$\times$1408 & 35.1 & 27.0 & 6.24 & 43.75 & 19.67 & 42.69 & 50.68 & 15.04 & 22.16 & 21.3 & 20.7 & 30.74 &  36.69 & 79.79 & 38.41 & 48.16 & 51.71 & 37.87 & 31.75\\
\bottomrule[1.5pt]
\end{tabular}
\end{adjustbox}
\label{table:sota_occ_eval}
\end{table*}
\vspace{6mm}

\begin{table*}[t]\tiny
  \centering
  \caption{Comparison of different paradigms on the nuScenes \texttt{val} set. $*$ initialized from a FCOS3D backbone. $\S$ with Class-balanced Grouping and Sampling (CBGS).}
  	\resizebox{1.0\linewidth}{!}{
    \begin{tabular}{l|cc|c|c|c|ccccc}
    \toprule[1.5pt]
    Methods & Backbone &Image Size         & \textbf{mAP}$\uparrow$ &\textbf{NDS}$\uparrow$ &\textbf{FPS}$\uparrow$ & mATE$\downarrow$  & mASE$\downarrow$   & mAOE$\downarrow$  & mAVE$\downarrow$  &  mAAE$\downarrow$    \\
    \midrule
    DETR3D$\S$\cite{DETR3D}    & R101* &900$\times$1600 & 30.3 & 37.4 & - & 86.0 & 27.8 & 43.7 & 96.7 & 23.5 \\
    PGD$\S$\cite{PGD}          & R101* &900$\times$1600 & 33.5 & 40.9 & - & 73.2 & 26.3 & 42.3 & 1.28 & 17.2 \\
    BEVDet$\S$\cite{huang2021bevdet}    & R101* &384$\times$1056 & 31.7 & 38.9 & - & 70.4 & 27.3 & 53.1 & 94.0 & 25.0 \\
    BEVDet$\S$\cite{huang2021bevdet}    & SwinT &512$\times$1408 & 34.9 & 41.7 & - & 63.7 & 26.9 & 49.0 & 91.4 & 26.8 \\
    PETR$\S$\cite{petr}        & R101\textcolor{white}{*} &512$\times$1408 & 35.7 & 42.1 & - & 71.0 & 27.0 & 49.0 & 88.5 & 22.4 \\
    PETR$\S$\cite{petr}        & R101* &512$\times$1408 & 36.6 & 44.1 & - & 71.7 & 26.7 & 41.2 & 83.4 & 19.0 \\
    PETR$\S$\cite{petr}        & SwinT &512$\times$1408 & 36.1 & 43.1 & - & 73.2 & 27.3 & 49.7 & 80.8 & 18.5 \\
    \midrule
    \bf{UltimateDO} & R101* &512$\times$1408 & 36.6 & 43.6 & 27.0 & 68.0 & 27.9 & 54.4 & 75.4 & 21.2 \\
    \bottomrule[1.5pt]
    \end{tabular}%
    }
  \label{tab:nus-val}%
\end{table*}%

\subsection{Experimental Setup}\label{sec:exp_setup}
\textbf{Benchmark.}
We conducted occupancy and 3D object detection using the Occ3D-nuScenes~\cite{tian2023occ3d} and nuScenes datasets~\cite{caesar2020nuscenes}, respectively.
The Occ3D-nuScenes dataset comprises 700 scenes for training and 150 scenes for validation. The dataset covers a spatial range of -40m to 40m along the X and Y axes, and -1m to 5.4m along the Z axis. The occupancy labels are defined using voxels with dimensions of $0.4m \times 0.4m \times 0.4m$ for 17 categories.
The nuScenes dataset consists of 1,000 distinct driving scenes, which are divided into three subsets: 700 training scenes, 150 validation scenes, and 150 testing scenes. Each driving scene contains 20 seconds of annotated perceptual data captured at a frequency of 2 Hz. The dataset encompasses 16 categories, including 6 background stuff categories and 10 foreground thing categories.
The data collection vehicle is equipped with one LiDAR, five radars, and six cameras, enabling a comprehensive surround view of the vehicle's environment.

\textbf{Evaluation metrics.} 
For occupancy prediction, the mean intersection-over-union (mIoU) over all classes is reported.
For 3D object detecion, we follow the official protocol to report the nuScense Score (NDS), mean Average Precision (mAP), along with five true positive metrics including mean Average Translation Error (mATE), mean Average Scale Error (mASE), mean Average Orientation Error (mAOE), mean Average Velocity Error (mAVE) and mean Average Attribute Error (mAAE).

\textbf{Training Details.}
For comprehensice comparison, we have conducted experiments with ResNet-50 and ResNet-101 as the backbone networks in our experiments.
The architecture of the detection branch is mainly following the setting in BEVDet~\cite{huang2021bevdet}, wherein the scope extent of nuScenes dataset is [-51.2m, -51.2m, -5m, 51.2m, 51.2m, 3m].
The occupancy prediction head refers from FlashOcc~\cite{yu2023flashocc}, and the scope extent of Occ3D-nuScenes is [-40m, -40m, -1m, 40m, 40m, 5.4m].
Given that occupancy prediction and 3D object detection are jointly trained in our framework, 
to encompass the range of the aforementioned two benchmarks, the scope range is defined as -51.2m to 51.2m in x/y dimension and -5m to 5.4m in height dimension.
The full-size BEV feature is forwarded to 3D object branch, whereas the center-cropped $100/128*S_{BEV} \times 100/128*S_{BEV}$ BEV feature is sent to occupancy branch, the $S_{BEV}$ is set to 128 in our experiments.
Unless otherwise stated, 
adamW optimizer with gradient clip is utilized, the learning rate is set to 2e-4.
The loss weight for cross-entropy loss in occupancy prediction $\lambda_{Occ}$ is set to 5.0 defaultly.
Only flip augmentation is utilized for BEV augmentation, while scale and rotate augmentations are not employed.
All experiments with a total batch size of 16 are trained for 24 epochs on 4 NVIDIA RTX3090 GPUs. 
Note that CBGS is not used for all experiments, and test augmentation methods are not utilized during the inference.

\subsection{Comparison with State-of-the-art Methods}\label{sec:sota_exp}
We compared the performance of our framework with other state-of-the-art methods on occupancy and 3D detection respectively. 

In terms of occupancy prediction, we directly compare our framework with popular existing approaches, i.e. MonoScene~\cite{cao2022monoscene}, TPVFormer~\cite{huang2023tri}, OccFormer~\cite{zhang2023occformer}, CTF-Occ~\cite{tian2023occ3d} and RenderOCC~\cite{pan2023renderocc}. Besides, we also extend the main-stream BEVFormer~\cite{li2022bevformer} to occupancy prediction follow setting in CTF-Occ~\cite{tian2023occ3d}. 
As listed in Table.~\ref{table:sota_occ_eval}, 
3D occupancy prediction performances on the Occ3D-nuScenes valuation dataset are listed. The results with ResNet-101 as backbone and $512\times1408$ as input size is evaluated, 
Our UltimateDO achieve 35.1 mIoU with 27.0 FPS. Note that the FPS is tested with the hole model including both detection and occupancy head.

\begin{figure*}
\centering
		\includegraphics[width=1.0\linewidth]{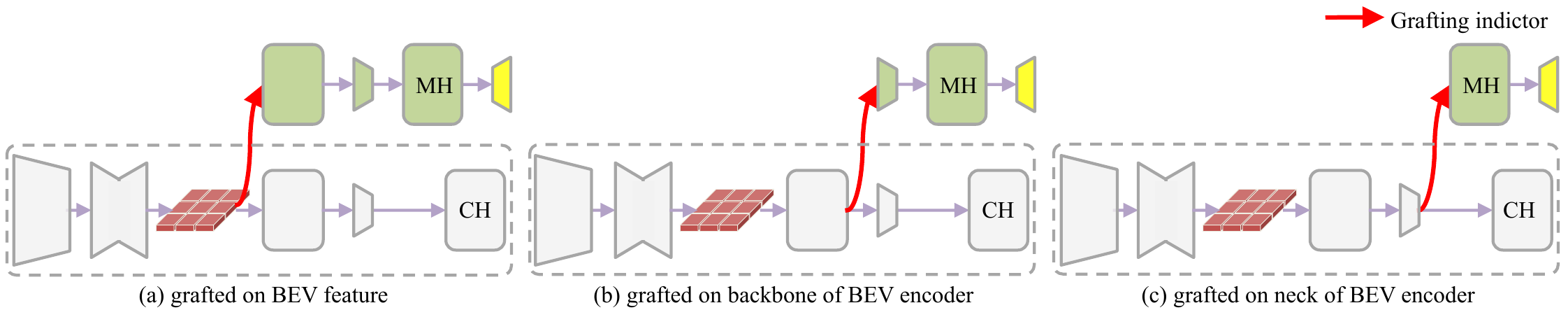}
	\caption{Illustration of occupancy branch grafted on (a) BEV feature; (b) backbone of BEV encoder; (c) neck of BEV encoder.
    The abbreviation ”MH” represents MLP head, and ”CH” stands for CenterPoint head.
    Icon interpretation descriptions follows Figure.~\ref{fig:fig1}.
    With the exception of the red line indicating the grafting location for occupancy branch, all other icons strictly adhere to the instructions depicted in Figure~\ref{fig:fig1}.
    }
	\label{fig:illustration_of_occupancy_branch_grafted_on}
\end{figure*}

In terms of 3D object detection, our UltimateDO, implemented without CBGS and without scale/rotate-augmentation for BEV feature, demonstrates substantial improvements in performance compared to competitors who utilize the CBGS strategy for data balance and employ scale/rotate-augmentation for robust BEV feature.
Specifically, our UltimateDO outperforms the origin BEVDet model with SwinT as the backbone by a margin of 1.7 mAP and 1.9 NDS. 
Additionally, when compared to the PETR model with SwinT as the backbone, our UltimateDO demonstrates superior performance with an improvement of 0.5 mAP and 0.5 NDS.
Even when a enhanced ResNet-101 backbone, initialized from the pre-trained FCOS3D backbone, is employed for PETR, our UltimateDO maintains the same mAP performance and only suffers a marginal decrease of 0.5 NDS. These results highlight the effectiveness of our proposed UltimateDO in achieving competitive performance in 3D object detection tasks.

\subsection{Ablation Study}\label{sec:exp_ablation}
We conduct ablative experiments to investigate the effect of each component in our framework. All the experiments are performed without CBGS strategy. 
Unless stated otherwise, we employ ResNet-50 as the backbone network with a input image resolution of 704 x 256. 
The default configuration did not include the temporal fusion module.


\textbf{Where to Graft Occupancy Branch.}\label{feature_decoupling}
The grafting location of the occupancy branch at an appropriate juncture is of paramount importance, as it influences: (1) the overall computational time and (2) the intricate interdependence and entanglement between detection and occupancy branches.
Figure.~\ref{fig:illustration_of_occupancy_branch_grafted_on} illustrates three progressive grafting schemes from shallow to deep.
As illustrated in Table.~\ref{tab:impaction_of_grafting_locations}, 
consider the setting when the occupancy branch is grafting on BEV feature and neck of BEV encoder,
both approaches demonstrate comparable performance. However, there is a notable increase in time consumption of 3.6ms.
Upon further examination of the comparison between all rows in Table ~\ref{tab:impaction_of_grafting_locations}, it is evident that the highest performance is attained when the grafting location is on the backbone of the BEV encoder. In comparison to the second competitor, this configuration yields a notable improvement of 0.3 mIoU, 0.4 mAP, and 0.4 NDS across all metrics.
It further demonstrates that appropriately joint training the detection and occupancy features helps to integrate knowledge from diverse scenes, consequently leading to a mutually enhancing performance.
However, the time-consumption is still 3ms larger compared to the setting when the grafting location is at the neck of BEV encoder. 
In conclusion, for large-scale offline inference in cloud-based settings, to achieve optimal perceptual accuracy, the configuration of grafting at backbone of BEV encoder is recommended. However, for deployment on edge devices with limited computational resources, we suggest adopting the more cost-effective configuration wherein the occupancy branch is graft on the neck of BEV encoder.

\begin{table}[htb] 
    \centering
    \begin{tabular}{l|c|c|c|c}
    \toprule[1.5pt]
     Grafting Location & mIoU & mAP & NDS & Time \\ 
     \midrule
     BEV feat. & 32.3 & 29.9 & 36.7 & 13.2ms \\
     Bac. of BEV enc. & 32.5 & 30.3 & 37.2 & 12.6ms \\
     Neck of BEV enc. & 32.2 & 29.7 & 36.8 & \textcolor{white}{0}9.1ms \\
    \bottomrule[1.5pt]
    \end{tabular}
    \caption{Impaction of Grafting Locations. 
    "Time" denotes the time-consumption tested on RTX3090 by tensorrt with fp16 precision. 
    "Bac." and "enc." short for backbone and encoder respectively. 
    }
    \label{tab:impaction_of_grafting_locations}
\end{table}

\textbf{Facilitation from Joint Training.}\label{unified_training}
As obervation in last graph that sharing the modules before the neck of BEV encoder (Figure.~\ref{fig:illustration_of_occupancy_branch_grafted_on} c) can facilitate mutual improvement of occupancy prediction and 3D object detection, with the optimal trade-off between precision and time-consumption. 
We conducted further ablation experiments to demonstrate the improvement achieved by joint training compared to separate training. 
As listed in Table.~\ref{tab:impaction_of_grafting_locations}, joint training resulted in an improvement of 2.3 mIoU, 1.8 NDS and 1.4 mAP over independent trained occupancy or 3D object detection task,
and only 1.1ms is consumed when occupancy branch is introduced to the detection module. 
The results demonstrate that fine-grained occupancy knowledge contributes to the construction of 3D semantic understanding within the model. Simultaneously, instance-level 3D bounding boxes aid in regulating the foreground space where objects are present. This reciprocal relationship between occupancy prediction and 3D object detection highlights their potential to mutually facilitate each other in enhancing overall performance.

\begin{table}[htb] 
    \centering
    \begin{tabular}{c|c|c|c|c|c}
    \toprule[1.5pt]
     Occ. & Det. & mIoU & NDS & mAP & Time \\ 
     \midrule
     \checkmark & & 29.9 & - & - & 8.2ms \\
     & \checkmark & - & 35.0 & 28.3 & 8.0ms \\
     \checkmark & \checkmark & 
     32.2{\fontsize{6}{14}\selectfont $\textcolor{red}{+2.3}$} &
     36.8{\fontsize{6}{14}\selectfont $\textcolor{red}{+1.8}$} &
     29.7{\fontsize{6}{14}\selectfont $\textcolor{red}{+1.4}$} & 9.1ms \\
    \bottomrule[1.5pt]
    \end{tabular}
    \caption{The improvement achieved by joint training compared to separate training.
    "Time" denotes the time-consumption tested on RTX3090 by tensorrt with fp16 precision. 
    "OCC." denotes occupancy task, "Det." indicates 3D object detection.
    }
    \label{tab:impaction_of_grafting_locations}
\end{table}

\textbf{Impact of Loss Weight $\lambda_{Occ}$ for Joint Training.}\label{loss_weight} 
Due to feature coupling, different tasks in multi-task training may interfere with each other. It is common practice to use weight coefficients to adjust the importance of different tasks.
We maintain the original detection loss weight to be 1.0, and experiment with different occupancy loss weight $\lambda_{Occ}$, specifically setting $\lambda_{Occ}$ to 1.0, 5.0 and 8.0 respectively.
The outcomes are presented in Table~\ref{tab:loss_weight}. It is evident that as the value of $\lambda_{Occ}$ increases, there is a noticeable decline in mIou, whereas mAP and NDS exhibit a tendency to rise.
For a compromise balance, the loss weight $\lambda_{Occ}$ is set to 5.0.

\begin{table}[htb] 
    \setlength{\tabcolsep}{6.5mm}
    \centering
    \begin{tabular}{c|c|c|c}
    \toprule[1.5pt]
     $\lambda_{Occ}$ & mIoU & NDS & mAP \\ 
     \midrule
     1.0 & 29.5 & 36.7 & 30.4 \\
     5.0 & 32.2 & 36.9 & 29.7 \\
     8.0 & 32.7 & 34.9 & 28.3 \\
    \bottomrule[1.5pt]
    \end{tabular}
    \caption{Impact of Loss Weight $\lambda_{Occ}$ for Joint Training.
    }
    \label{tab:loss_weight}
\end{table}

\textbf{Training without pretraining}\label{Benifit from training without pretraining}

\begin{table}[htb] 
    \setlength{\tabcolsep}{2.8mm}
    \centering
    \begin{tabular}{c|c|c|c|c|c}
    \toprule[1.5pt]
     Det. Init. & Occ. & Det. & mIoU & NDS & mAP \\ 
     \midrule
     & \checkmark & & 29.9 & - & - \\
     & & \checkmark & - & 35.0 & 28.3 \\
     \checkmark & \checkmark & & 32.0 & - & - \\
     & \checkmark & \checkmark & 
     32.2 &
     36.8 &
     29.7 \\
    \bottomrule[1.5pt]
    \end{tabular}
    \caption{Benifit from joint training compared to separate training with pretrained detection.
    "Det. Init." represents that the model is initialized from the weight pretrained on BEVDet (second row in current Table).
    "OCC." denotes occupancy task, "Det." indicates 3D object detection.
    }
    \label{tab:impaction_of_grafting_locations}
\end{table}

\textbf{Influence of BEV Augmentations.}\label{Influence_of_Augmentation}
In general, BEV Augmentations typically involve operations such as flip, rotation, and scaling. Previous studies have shown that flip augmentation is beneficial for both occupancy and 3D object detection. However, scaling augmentation can introduce truncation errors, leading to a degradation in performance. For this reason, it is not utilized in the BEVDetOcc, RenderOcc, and FBOcc methods.
Additionally, while rotation augmentation is employed in FBOcc, it is not used in BEVDetOcc and RenderOcc. Therefore, we aim to investigate the impact of rotation augmentation on joint training. According to the results presented in Table.~\ref{tab:influence_of_bev_augmentation}, 
it is observed that rotation augmentation makes no difference on mIoU but leads to inferior NDS scores, which contradicts the notion that BEV augmentation improves 3D object detection performance.
Consequently, we have decided to retain only the flip augmentation as the BEV augmentation method for our joint training approach.

\begin{table}[htb] 
    \setlength{\tabcolsep}{5.0mm}
    \centering
    \begin{tabular}{l|c|c|c}
    \toprule[1.5pt]
     BEVAug & mIoU & NDS & mAP \\ 
     \midrule
     Flip & 32.2 & 36.9 & 29.7 \\
     Flip+Rotation & 32.1 & 36.2 & 30.0 \\
    \bottomrule[1.5pt]
    \end{tabular}
    \caption{Influence of BEV Augmentation. 
    }
    \label{tab:influence_of_bev_augmentation}
\end{table}

\section{Conclusion}
To efficiently achieve occupancy prediction and 3D object detection, 
we marry the occupancy prediction head in FlashOcc to the BEV neck of BEVDet, 
resulting in a unified framework called UltimateDO. 
This integration allows UltimateDO to be easily deployed on various computing chips while maintaining high precision with minimal time consumption.
Notably, UltimateDO does not rely on 3D convolution or transformer modules, and the occupancy prediction branch and detection branch can mutually benefit each other. 
Besides, merely 1.1ms computation for the grafted occupancy head. 
We believe that UltimateDO can serve as a robust baseline for the research community, and we are committed to expanding its application to other perception pipelines in autonomous driving systems.

{\small
\bibliographystyle{ieee_fullname}
\bibliography{main}
}


\end{document}